# Mean Nyström Embeddings for Adaptive Compressive Learning

**Antoine Chatalic**
DIBRIS & MaLGA,
Università di Genova

**Luigi Carratino**
DIBRIS & MaLGA,
Università di Genova

**Ernesto De Vito**
DIMA & MaLGA,
Università di Genova

**Lorenzo Rosasco**
DIBRIS & MaLGA,
Università di Genova
CBMM, MIT, IIT

## Abstract

Compressive learning is an approach to efficient large scale learning based on compressing an entire dataset to a single mean embedding (the sketch), i.e. a vector of generalized moments. The learning task is then approximately solved as an inverse problem using an adapted parametric model. Previous works in this context have focused on sketches obtained by averaging random features, that while universal can be poorly adapted to the problem at hand. In this paper, we propose and study the idea of performing sketching based on data-dependent Nyström approximation. From a theoretical perspective we prove that the excess risk can be controlled under a geometric assumption relating the parametric model used to learn from the sketch and the covariance operator associated to the task at hand. Empirically, we show for k-means clustering and Gaussian modeling that for a fixed sketch size, Nyström sketches indeed outperform those built with random features.

## 1 INTRODUCTION

Various approaches have been proposed to scale standard machine learning techniques to large datasets. For instance, dimensionality reduction techniques help to cut down the cost of processing each sample, coresets can be used to reduce the dataset size (Feldman 2020), and low-rank or structured approximations techniques are helpful when working with kernel methods to avoid building the full kernel matrix (Rahimi et al. 2008; Teneva et al. 2016). In this paper we focus on compressive learning (Gribonval et al. 2021a), an approach which consists in compressing the whole dataset down to a single vector of generalized moments, called the sketch. An approximate solution to the learning task can then be inferred from this sketch by fitting a parametric model of interest, without using the initial data. In fact, any subsequent learning operation solely based on the sketch can be performed with time and space complexities wich are independent of the number of samples in the collection. Moreover, although the sketch needs to be somehow adapted to the learning task to solve, once computed it can be reused multiple times. This framework has already been successfully applied on a few unsupervised learning tasks such as k-means clustering and Gaussian modeling (Keriven et al. 2017a,b). The core idea of this approach is that distributions can be represented via mean embeddings in a feature space (Muandet et al. 2017). Kernel mean embeddings are an example of such a representation, and the associated mapping is known to be injective when the kernel is characteristic (Sriperumbudur et al. 2010). However, while such embeddings typically belong to large- or infinite-dimensional Hilbert spaces, compressive learning exploits the fact that they can be further reduced to compact finite-dimensional vectors while approximately preserving the geometry between embeddings for a family of distributions of interest.

Previous works focused on data-independent approximation schemes, such as sketches obtained by averaging random features (Bourrier et al. 2013). In this work, we suggest to use instead the mean embeddings associated with a Nyström approximation (Williams et al. 2001). The latter is data-dependent, i.e. the approximation is adaptive to the dataset to sketch. As a consequence we expect to potentially be able to reach a desired accuracy using a smaller sketch size compared to when using random features. Indeed we observe this behavior experimentally for k-means clustering and Gaussian modeling. From a theoretical perspective, the adaptive nature of the sketching operator makes the analysis different than for random features as the data distribution must now somehow be compatible with the parametric model used to learn from the sketch. We propose a way to characterize this compatibility and derive a bound on the learning excess risk under this assumption. Our result covers the settings where the points used to design the Nyström approximation are sampled from the dataset either uniformly or using approximate leverage scores.

The plan of the paper is as follows. We introduce the compressive learning in Section 2, and introduce the Nyström sketches in Section 3. We provide theoretical results on the control of the excess risk in Section 4, and an experimental validation in Section 5.

## 2 FROM EMPIRICAL RISK MINIMIZATION TO COMPRESSIVE LEARNING

We introduce the statistical learning setting in Section 2.1, and compressive learning in Section 2.2.

### 2.1 Statistical machine learning

Let $(\mathcal{X}, \mathcal{B}, \pi)$ be a probability space where $\mathcal{X}$ is a locally compact second countable topological space, $\mathcal{B}$ the Borel $\sigma$-algebra and $\pi$ a probability distribution to be interpreted as a data distribution over $\mathcal{X}$. We consider an hypothesis space $H$ and a loss function $l : \mathcal{X} \times H \to \mathbb{R}$, which naturally defines a risk function $\mathcal{R} : \mathcal{P}(\mathcal{X}) \times H \to \mathbb{R}, \pi \mapsto \mathbf{E}_{\mathbf{x} \sim \pi} l(\mathbf{x}, h)$, where $\mathcal{P}(\mathcal{X})$ denotes the space of probability distributions over $\mathcal{X}$. We are interested in finding an hypothesis $\hat{h} \in H$ minimizing the so-called excess risk

$$\mathrm{ER}(\pi, \hat{h}) \triangleq \mathcal{R}(\pi, \hat{h}) - \inf_{h \in H} \mathcal{R}(\pi, h). \quad (1)$$

This objective is intractable as $\pi$ is unknown. Yet, given a dataset $\mathbf{X} = \{\mathbf{x}_1, ..., \mathbf{x}_n\}$ one can define the empirical probability distribution $\pi_n \triangleq \frac{1}{n} \sum_{i=1}^n \delta(\mathbf{x}_i)$, where $\delta$ denotes the Dirac delta, and solve instead

$$\inf_{h \in H} \mathcal{R}(\pi_n, h), \quad (2)$$

which is known as empirical risk minimization.

**Learning tasks** In this paper, following previous works on compressive learning (Gribonval et al. 2021b, Section 3 and 4), we will mainly focus on two unsupervised learning tasks, namely clustering and Gaussian modeling. We detail the hypothesis spaces considered for these two problems.

**Example 1** (Clustering)**.** *We consider $\mathcal{X} = \mathbb{R}^d$ and $H = \{h = (\mathbf{h}_1, ..., \mathbf{h}_k) \in \mathcal{X}^k\}$. The k-means and k-medians clustering problems consists in minimizing the risk induced by the loss $l(\mathbf{x}, h) = \min_{1 \le i \le k} \|\mathbf{x} - \mathbf{h}_i\|_2^p$, taking respectively $p = 2$ and $p = 1$.*

**Example 2** (Gaussian modeling)**.** *For $\mathcal{X} = \mathbb{R}^d$ and a known invertible covariance matrix $\mathbf{\Gamma} \in \mathbb{R}^{d \times d}$, $H$ is a family of means and weights $h = (\boldsymbol{\mu}_1, ..., \boldsymbol{\mu}_k, \alpha_1, ..., \alpha_k)$ of the Gaussian mixture model $\pi_h = \sum_{i=1}^k \alpha_i \mathcal{N}(\boldsymbol{\mu}_i, \mathbf{\Gamma})$, where for each $i \in [1, k]$ $\boldsymbol{\mu}_i \in \mathcal{X}$, and the weights $\alpha_1, ..., \alpha_k$ are positive and sum to one. The loss function is the negative log-likelihood $l(\mathbf{x}, h) = -\log \pi_h(\mathbf{x})$.*

Although the covariance matrix is fixed in Example 2 in order to simplify the theoretical analysis, the experiments conducted in Section 5 consist in learning different (diagonal) covariance matrices for the $k$ components of the mixture. In the following, we will add additional restrictions on the hypothesis spaces considered for both clustering and Gaussian modeling, but we stick for now to these definitions for conciseness.

### 2.2 Compressive learning with moments

Solving the empirical risk minimization problem (2) typically requires going multiple times through the dataset, which can be prohibitive for large collections. One way to avoid this problem is to replace the empirical risk $\mathcal{R}(\pi_n, \cdot)$, which explicitly depends on all the data samples, by a proxy function $\tilde{\mathcal{R}}(\tilde{\mathbf{s}}, \cdot)$ where $\tilde{\mathbf{s}} \in \mathbb{R}^m$ is a small *sketch* summarizing the data collection and computed in one pass over the data. The approach thus consists of two steps:

1. the whole dataset $\mathbf{X} \in \mathbb{R}^{d \times n}$ is compressed down to a single sketch $\tilde{\mathbf{s}} \in \mathbb{R}^m$ (sketching step);
2. an approximate solution to the learning problem (2) is recovered from the sketch, *without using the original data* (learning step).

We stress that the above procedure is somehow different from many classical approaches for large scale learning where the algorithm strongly depends on the task to be solved. We now detail how these two steps are performed.

**Sketching step** In this work, we consider sketches made of (generalized) moments of the data, i.e. that can be expressed as

$$\tilde{\mathbf{s}} \triangleq \frac{1}{n} \sum_{i=1}^n \Phi_m(\mathbf{x}_i) \quad \text{where} \quad \Phi_m : \mathcal{X} \to \mathbb{R}^m \quad (3)$$

is a feature map taking values in $\mathbb{R}^m$. In the following, it will be useful to think of the dataset $\mathbf{X}$ through its empirical distribution $\pi_n$, and we thus rewrite $\tilde{\mathbf{s}} = \mathcal{A}_m(\pi_n)$ where $\mathcal{A}_m$ is the sketching operator

$$\mathcal{A}_m : \mathcal{P}(\mathcal{X}) \to \mathbb{R}^m \quad \mathcal{A}_m(\pi) = \int_{\mathcal{X}} \Phi_m(\mathbf{x}) \, d\pi(\mathbf{x}) \quad (4)$$

whose properties are reviewed in Appendix B. Naturally, finding a feature map $\Phi_m$ such that the sketch (3) summarizes all the information required to solve the desired learning task is highly challenging and not always possible. Nonetheless, some specific learning tasks are known to be compatible with this approach; this is in particular the case of principal component analysis (PCA), k-means and Gaussian modeling, for which we have both empirically working algorithms and theoretical guarantees on the excess risk of the recovered solution.

**Example 3** (PCA with centered data). *For $\mathcal{X} = \mathbb{R}^d$, the PCA solution depends only on the data covariance matrix. Thus, assuming centered data and denoting* vec *the vectorization operation, the feature map $\Phi_m(\mathbf{x}) = \text{vec}(\mathbf{x}\mathbf{x}^T)$ with $m = d^2$ contains all the information required to solve the problem.*

**Example 4** (Random Fourier features). *When $\mathcal{X} = \mathbb{R}^d$ and the feature map takes the form*

$$\Phi(\mathbf{x}) = [\cos(\mathbf{\Omega}^T \mathbf{x}), \sin(\mathbf{\Omega}^T \mathbf{x})] \in \mathbb{R}^m \quad (5)$$

*where $m = 2m'$ and $\mathbf{\Omega} \in \mathbb{R}^{d \times m'}$ is a random matrix with i.i.d. normal entries and the* cos *and* sin *functions are applied pointwise, we obtain a mean vector of random Fourier features (Rahimi et al. 2008). Such embeddings have successfully been used to solve the clustering and Gaussian modeling tasks in a compressive manner (Gribonval et al. 2021a,b). When $m \to \infty$, the inner-product $\langle \Phi(\mathbf{x}), \Phi(\mathbf{y}) \rangle$ approximates with growing accuracy the Gaussian kernel $\kappa(\mathbf{x}, \mathbf{y}) = \exp(-\frac{1}{2} \|\mathbf{x} - \mathbf{y}\|^2)$.*

Computing a mean sketch of the form (3) has many advantages. Provided that $m \ll nd$, storing and manipulating the sketch is much more efficient than manipulating the raw data. The time complexity of sketching is linear in the number of samples $n$ in the collection, and all subsequent operations performed on the sketch have a complexity which *does not depend on $n$*. Moreover, as the original data can be discarded once the sketch is computed, sketching is also an interesting tool for privacy preservation (Chatalic et al. 2021).

**Learning step** Previous works on random Fourier sketches (Gribonval et al. 2021a,b) showed that multiple learning tasks can be tackled as moment-matching problems of the form

$$\hat{\pi} \triangleq \arg\min_{p \in \mathfrak{S}} \|\mathcal{A}_m(p) - \mathcal{A}_m(\pi_n)\|_2 \quad (6)$$

where $\mathfrak{S}$ is a parametric family of probability distributions adapted to the learning problem and plays the role of a regularizer. This problem is typically non-convex, but multiple heuristics have been developed using techniques such as hard thresholding (Bourrier et al. 2013), approximate message passing (Byrne et al. 2019) or orthogonal matching pursuit (Keriven et al. 2017a). The latter algorithm (called CL-OMP) is a generic approach usable when $\mathfrak{S}$ is a mixture model, and consists in iteratively building the desired mixture $\hat{\pi}$ by alternating between greedily adding a new atomic distribution minimizing the residual of the objective cost, and globally optimizing all the parameters and weights of the mixture. When applied to the clustering problem, it can be interpreted as a variation of the Frank-Wolfe algorithm.

The problem (6) can be viewed an inverse problem: if the samples $\mathbf{X}$ are drawn i.i.d. from $\pi$ and $n$ is large enough, we have $\mathcal{A}_m(\pi_n) \approx \mathcal{A}_m(\pi)$ and one can think of $\mathcal{A}_m(\pi_n)$ as a noisy observation of the distribution $\pi$ via the linear operator $\mathcal{A}_m$. Remember that we want in the end to solve (2), and thus we recover an hypothesis $\hat{h} \in H$ by solving

$$\hat{h} \triangleq \inf_{h \in H} \mathcal{R}(\hat{\pi}, h) \quad (7)$$

This step is typically costless, given that the structure of $\mathfrak{S}$ will most often be closely related to the structure of the hypothesis space $H$, i.e. an optimal hypothesis $\hat{h}$ can directly be recovered from the probability distribution $\hat{\pi}$ in the examples that we consider. We now provide two examples of model sets coming from Gribonval et al. (2021b, Section 3&4), which implicitly depend on the chosen hypothesis space $H$.

**Example 5** (k-means clustering). *We consider*

$$\mathfrak{S} = \bigcup_{h=(\mathbf{c}_1, ..., \mathbf{c}_k) \in H^{\text{Cl}}} \left\{ \sum_{i=1}^k \alpha_i \delta(\mathbf{c}_i) \,\middle|\, \sum_{i=1}^k \alpha_i = 1, \alpha_i \geq 0 \right\}.$$

*An hypothesis satisfying (7) can be recovered from $\hat{\pi} \in \mathfrak{S}$ by keeping the locations $(\mathbf{c}_1, ..., \mathbf{c}_k)$ of the $k$ Diracs and dropping the weights.*

**Example 6** (Gaussian modeling). *Following Gribonval et al. (2021b, Sc.4), we use $\mathfrak{S} = \{\pi_h : h \in H^{\text{GMM}}\}$ where $\pi_h$ is the Gaussian mixture with means and weights $h = (\boldsymbol{\mu}_1, ..., \boldsymbol{\mu}_k, \alpha_1, ..., \alpha_k)$, see (2).*

**Separation assumption** Although one can define the model sets from Examples 5 and 6 using the hypothesis space from Examples 1 and 2, it turns out that additional restrictions are required to carry out the theoretical analysis. We will thus rather consider the hypothesis space $H$ to be

- the set of tuples $(\mathbf{c}_1, ..., \mathbf{c}_k)$ s.t. $\min_{i \neq j} \|\mathbf{c}_i - \mathbf{c}_j\| \geq 2\varepsilon$ and $\max_l \|\mathbf{c}_l\| \leq R$ for clustering;
- the set of centers and weights $(\mathbf{c}_1, ..., \mathbf{c}_k, \alpha_1, ..., \alpha_k)$ s.t. $\min_{i \neq j} \|\mathbf{c}_i - \mathbf{c}_j\|_{\mathbf{\Gamma}} \geq \varepsilon$, $\max_l \|\mathbf{c}_l\|_{\mathbf{\Gamma}} \leq R$, $\forall i \alpha_i \geq 0$, and $\sum_{1 \leq i \leq k} \alpha_i = 1$ for Gaussian modeling, where $\|\mathbf{x}\|_{\mathbf{\Gamma}} = (\mathbf{x}^T \mathbf{\Gamma}^{-1} \mathbf{x})^{1/2}$ denotes the Mahalanobis distance.

These two definitions add a separation assumption between the Diracs (for clustering) or the centers of the components (Gaussian modeling), which is known to be necessary for compressive learning in this setting (Gribonval et al. 2021b, Section 3.2).

## 3 MEAN NYSTRÖM FEATURES

Random Fourier features are easily computable and have been used in many contexts. They are generic in the sense that they are data-independent: the distribution of the matrix $\mathbf{\Omega}$ in (5) does not depend on the data to sketch. In this work, we advocate using a *data-dependent* feature map that we now introduce.

### 3.1 The Nyström feature map

The Nyström feature map derives from a similarity metric $\kappa : \mathcal{X} \times \mathcal{X} \to \mathbb{R}$. We assume in the following that $\kappa$ is a positive definite kernel, i.e. $\kappa$ is symmetric and for any choice of $(\mathbf{x}_i, ..., \mathbf{x}_j) \in \mathcal{X}^m$ the $n \times n$-matrix $[\kappa(\mathbf{x}_i, \mathbf{x}_j)]_{1 \leq i \leq m, 1 \leq j \leq m}$ is positive semi-definite. We select a set of landmark points $\tilde{\mathbf{X}} = (\tilde{\mathbf{x}}_1, ..., \tilde{\mathbf{x}}_m)$ which we want to be "representative" of the data set, i.e. for instance drawn from $\pi$ (or in practice sampled from the dataset). We denote $\mathbf{K}_m$ the associated $m \times m$ kernel matrix with entries $(\mathbf{K}_m)_{ij} = \kappa(\tilde{\mathbf{x}}_i, \tilde{\mathbf{x}}_j)$, which by assumption is symmetric and positive definite, so that $\text{Im}(\mathbf{K}_m) = \ker(\mathbf{K}_m)^\perp$. We denote by $\mathbf{K}_m^\dagger$ the pseudo-inverse of $\mathbf{K}_m$, and if $\mathbf{K}_m$ is invertible then $\mathbf{K}_m^\dagger = \mathbf{K}_m^{-1}$. Since $\mathbf{K}_m^\dagger$ is symmetric and positive definite, its square-root is well defined and we denote it by $\mathbf{K}_m^{-1/2}$ with a slight abuse of notation. Following Williams et al. (2001), we define the feature map as

$$\Phi_m(\mathbf{x}) \triangleq \mathbf{K}_m^{-1/2} \begin{bmatrix} \kappa(\tilde{\mathbf{x}}_1, \mathbf{x}) \\ \vdots \\ \kappa(\tilde{\mathbf{x}}_m, \mathbf{x}) \end{bmatrix} \in \mathbb{R}^m \qquad (8)$$

and the associated sketching operator as in (4).

The intuition here is that one would like $\langle \Phi_m(\mathbf{x}), \Phi_m(\mathbf{y}) \rangle$ to approximates well $\kappa(\mathbf{x}, \mathbf{y})$, maybe not uniformly on $\mathcal{X}^2$ but at least when $\mathbf{x}$ and $\mathbf{y}$ are similar to the landmarks $\tilde{\mathbf{X}}$ (which for many choices of kernel means when $\mathbf{x}$ and $\mathbf{y}$ are located where the mass of $\pi$ is concentrated). The factor $\mathbf{K}_m^{-1/2}$ in (8) should be interpreted as a geometric corrective factor and we will see in Section 3.3 where it comes from. In particular with this normalization we have $\langle \Phi_m(\tilde{\mathbf{x}}_i), \Phi_m(\tilde{\mathbf{x}}_j) \rangle = \kappa(\tilde{\mathbf{x}}_i, \tilde{\mathbf{x}}_j)$ for any pair of landmarks $\tilde{\mathbf{x}}_i, \tilde{\mathbf{x}}_j$.

**Learning from the sketch** In order to tackle the inverse problem (6) using first-order methods, one needs a closed form expression of the gradient of the objective function with respect to some parametrization of the model set $\mathfrak{S}$. We derive these expressions

---

**Algorithm 3.1:** Nyström Compressive Learning

**Input:** Dataset $\mathbf{X}$, kernel $\kappa$, model set $\mathfrak{S}$
**Output:** An hypothesis $\hat{h} \in H$

1. If needed, learn the kernel parameters using a small uniform i.i.d. subsample of $\mathbf{X}$;
2. If using leverage scores sampling, estimate the leverage scores (9) from $\mathbf{X}$ using a fast heuristic;
3. Sample $m$ landmarks $(\tilde{\mathbf{x}}_i)_{1 \leq i \leq m}$ from $\mathbf{X}$ (cf. discussion in Section 3.2);
4. Compute the sketch $\tilde{\mathbf{s}} = \frac{1}{n} \sum_{i=1}^n \Phi_m(\mathbf{x}_i)$ where $\Phi$ is defined from the landmarks in (8) ;
5. Find an approximate solution $\hat{\pi}$ of $\min_{p \in \mathfrak{S}} \|\mathcal{A}_m(p) - \tilde{\mathbf{s}}\|_2$ using an heuristic such as CL-OMP (Keriven et al. 2017a) ;
6. Return $\hat{h} \in \inf_{h \in H} \mathcal{R}(\hat{\pi}, h)$ (usually costs $O(1)$);

---

in Appendix A for the feature map (8) for the tasks of k-means clustering and Gaussian modeling using the parametrizations of Examples 5 and 6.

**Complexities** Computing $\mathbf{K}_m$ has a time-complexity of $\Theta(m^2 d)$ assuming that a kernel evaluation takes $\Theta(d)$ operations, and inverting $\mathbf{K}_m$ takes $\Theta(m^3)$. After that, the evaluation of $\Phi_m$ takes $\Theta(m^2 + md)$. Note that structured landmark matrices have been proposed to speed-up computations (Si et al. 2016) and could be used here to some extent.

We provide in Algorithm 3.1 a summary of the whole learning procedure, including the sampling step which we now detail more precisely.

### 3.2 Sampling schemes

Naturally, the projected features (8) considered in this section require to carefully select the landmark points $\tilde{\mathbf{X}}$. In this paper, we will always sample $\tilde{\mathbf{X}}$ from the empirical data $\mathbf{X} = (\mathbf{x}_1, ..., \mathbf{x}_n)$, and consider three different sampling schemes.

- **Uniform** The first considered scheme is *uniform sampling*, where the set $\tilde{\mathbf{X}}$ is sampled uniformly at random among all possible subsets of $\mathbf{X}$ of cardinality $m$.
- **Approximate leverage score (ALS)** The landmarks are sampled according to the approximate leverage scores of $\mathbf{X}$ (Alaoui et al. 2015). Let $\mathbf{K}_n \in \mathbb{R}^{n \times n}$ be the (full) kernel matrix. For $\lambda > 0$, the leverage scores of the set $\mathbf{X}$ are defined as

$$\ell(\lambda, i) = \left(\mathbf{K}_n (\mathbf{K}_n + \lambda n I)^{-1}\right)_{ii}, \quad \forall i \in [n]. \qquad (9)$$

Since computing these exact leverage scores can be prohibitive, approximate variants can be considered. Let $\delta \in (0, 1]$, $\lambda_0 > 0$ and $z \in [1, \infty)$.

Then, a sequence $(\hat{\ell}(\lambda, i))_{i \in [n]}$ consists of $(z, \lambda_0, \delta)$-ALS of $\mathbf{X}$ if it satisfies w.p. at least $1 - \delta$

$$\frac{1}{z} \ell(\lambda, i) \leq \hat{\ell}(\lambda, i) \leq z\, \ell(\lambda, i) \quad \forall \lambda \geq \lambda_0, \forall i \in [n]. \tag{10}$$

Different algorithms have been proposed to compute such approximations. In this work we use BLESS (Rudi et al. 2018) which uses a coarse to fine strategy and has a computational cost which is negligible compared to other operations. After computing the values $\hat{\ell}(\lambda, i)$, the landmarks are obtained sampling from $\mathbf{X}$ proportionally to $\hat{\ell}(\lambda, i)$.

- **Greedy diversity sampling** The third sampling scheme is a *greedy method* that aims at selecting the most diverse landmarks. The algorithm promotes large principal angles between landmarks by iteratively selecting the points according to their Schur complement (Carratino et al. 2021; Chen et al. 2018). In more details, let $\tilde{\mathbf{x}}_1 = \arg\max_{\mathbf{x} \in \mathbf{X}} \kappa(\mathbf{x}, \mathbf{x})$, then the $t$-th landmark $\tilde{\mathbf{x}}_t$ is selected as

$$\tilde{\mathbf{x}}_t \triangleq \underset{\mathbf{x} \in X / \{\tilde{\mathbf{x}}_1, \ldots \tilde{\mathbf{x}}_{t-1}\}}{\arg\max} \kappa(\mathbf{x}, \mathbf{x}) - \varphi_{t-1}(\mathbf{x})^\top \mathbf{K}_{t-1}^{-1} \varphi_{t-1}(\mathbf{x})$$

where $\varphi_{t-1}(\mathbf{x}) \triangleq [\kappa(\mathbf{x}, \tilde{\mathbf{x}}_1), \ldots, \kappa(\mathbf{x}, \tilde{\mathbf{x}}_{t-1})]$ and $\mathbf{K}_{t-1} \in \mathbb{R}^{t-1 \times t-1}$ is the kernel matrix of the already selected landmarks.

### 3.3 Nyström features are projected features

As we assumed that the function $\kappa$ used to build the Nyström feature map (8) is a positive definite kernel, there exists a Hilbert space $\mathcal{F}$ with inner-product $\langle \cdot, \cdot \rangle_\mathcal{F}$ and a feature map $\Phi : \mathcal{X} \to \mathcal{F}$ such that $\kappa(\mathbf{x}, \mathbf{y}) = \langle \Phi(\mathbf{x}), \Phi(\mathbf{y}) \rangle_\mathcal{F}$ for any $\mathbf{x}, \mathbf{y} \in \mathcal{X}^2$ (Steinwart et al. 2008, Theorem 4.16). The canonical choice is to set $\mathcal{F}$ to be the reproducing kernel Hilbert space uniquely defined by $\kappa$ and to define $\Phi(x) = \kappa(\cdot, x)$ for all $x \in \mathcal{X}$ (Steinwart et al. 2008). However multiple feature maps can be associated to the same kernel and in many applications there might exist more natural choices. For example, if $\mathcal{X}$ is a subset of $\mathbb{R}^d$ and $\kappa(\mathbf{x}, \mathbf{x}') = \mathbf{x}^T \mathbf{x}'$ is the linear kernel, then one can choose $\mathcal{F} = \mathbb{R}^d$ and $\Phi(\mathbf{x}) = \mathbf{x}$.

We now define $\mathcal{A}(p) \triangleq \mathbf{E}_{\mathbf{x} \sim p} \Phi(\mathbf{x}) \in \mathcal{F}$ the mean embedding of $p \in \mathcal{P}(\mathcal{X})$.

**Example 7.** *When $\mathcal{F}$ is a reproducing kernel Hilbert space (RKHS) and $\Phi$ the associated canonical feature map, the sketch $\mathcal{A}(\pi)$ can be interpreted as a kernel mean embedding (Muandet et al. 2017). When $\mu = p - q$ is a difference of two probability distributions, $d(p, q) \triangleq \|\mathcal{A}(\mu)\|_\mathcal{F}$ corresponds to the maximum mean discrepancy between $p$ and $q$, and is known to be a metric (i.e. the mean embedding is injective) iff the kernel $\kappa$ is characteristic (Sriperumbudur et al. 2010).*

We started in Section 3.1 with the definition the feature map $\Phi_m$ taking values in $\mathbb{R}^m$ because this is the one that is used in practice for efficient computations. Yet, from a theoretical perspective it is interesting to see that the feature map $\Phi_m$ is directly related to the projection of $\Phi$ onto the finite-dimensional subspace $\mathcal{F}_m = \text{span}\{\Phi(\tilde{\mathbf{x}}_1), \ldots, \Phi(\tilde{\mathbf{x}}_m)\}$ of $\mathcal{F}$. To formalize this statement we denote by $P_m$ the orthogonal projection onto $\mathcal{F}_m$.

**Lemma 3.1:** *There exists a bounded operator $U : \mathbb{R}^m \to \mathcal{F}$ satisfying $\ker(U) = \ker(\mathbf{K}_m)$ and $\|U\mathbf{c}\|_\mathcal{F} = \|\mathbf{c}\|$ for any $\mathbf{c} \in \ker(U)^\perp$ (i.e. $U$ is an isometry from $\ker(\mathbf{K}_m)^\perp$ onto $\mathcal{F}_m$) such that $\forall \mathbf{x}, \mathbf{y} \in \mathcal{X}$*

$$U\Phi_m = P_m \Phi \tag{11a}$$
$$U\mathcal{A}_m = P_m \mathcal{A} \tag{11b}$$
$$\langle \Phi_m(\mathbf{x}), \Phi_m(\mathbf{y}) \rangle = \langle P_m \Phi(\mathbf{x}), P_m \Phi(\mathbf{y}) \rangle_\mathcal{F}. \tag{11c}$$

This relation justifies in particular the choice of the normalization factor $\mathbf{K}_m^{-1/2}$ in (8).

## 4 THEORETICAL ANALYSIS

We introduce the setting in Section 4.1, state our main result in Section 4.2, and give an idea of the proof in Section 4.3.

### 4.1 Setting and assumptions

We assume that $\mathcal{F}$ is a separable Hilbert space with inner-product $\langle \cdot, \cdot \rangle_\mathcal{F}$ and norm $\|\cdot\|_\mathcal{F}$. We denote $\mathcal{L}(\mathcal{F})$ the set of bounded linear operators on $\mathcal{F}$ endowed with the operator norm $\|\cdot\|_{\mathcal{L}(\mathcal{F})}$, and $\kappa : \mathbf{x}, \mathbf{y} \mapsto \langle \Phi(\mathbf{x}), \Phi(\mathbf{y}) \rangle_\mathcal{F}$ the positive definite kernel associated with the feature map $\Phi$.

**Assumption 1.** *For every $\mathbf{x} \in \mathcal{X}$, $\|\Phi(\mathbf{x})\|_\mathcal{F} \leq K$ and $\Phi$ is measurable.*

A direct consequence of Assumption 1 is that for any probability distribution $p \in \mathcal{P}(\mathcal{X})$, $\Phi(\cdot)$ is $p$-integrable and the mean embedding $\mathcal{A}$ introduced in the previous section is well defined. We extend its definition to any finite signed measure $\mu$ in Appendix B.

**Integral operator** We define the (uncentered) covariance operator $\Sigma : \mathcal{F} \to \mathcal{F}$ as

$$\Sigma = \int \Phi(\mathbf{x}) \otimes_\mathcal{F} \Phi(\mathbf{x}) d\pi(\mathbf{x}),$$

where $\Phi(\mathbf{x}) \otimes_\mathcal{F} \Phi(\mathbf{x}) : f \mapsto \langle f, \Phi(\mathbf{x}) \rangle_\mathcal{F} \Phi(\mathbf{x})$ is a rank one operator and $\Sigma$ is a positive trace class operator on $\mathcal{F}$, see Appendix B. For any $f \in \mathcal{F}$ and $\lambda > 0$ we define $\mathcal{N}_f(\lambda) \triangleq \langle f, (\Sigma + \lambda I)^{-1} f \rangle_\mathcal{F}$ and, with slight abuse of notation, we write $\mathcal{N}_x(\lambda) \triangleq \mathcal{N}_{\Phi(\mathbf{x})}(\lambda)$ for all $\mathbf{x} \in \mathcal{X}$. We denote $\mathcal{N}(\lambda) \triangleq \mathbf{E}_x \mathcal{N}_x(\lambda) = \text{tr}(\Sigma(\Sigma + \lambda I)^{-1})$, which is known as the effective dimension or degrees of freedom. We also let $\mathcal{N}_\infty(\lambda) \triangleq \sup_x \mathcal{N}_x(\lambda)$, and it is easy to see that $\mathcal{N}_\infty(\lambda) \leq K^2/\lambda < \infty$ for any $\lambda > 0$ under Assumption 1.

**Assumption on the model** Given that the feature map (8) used for sketching is data-dependent, and considering that we recover $\hat{\pi}$ from the empirical sketch by solving the inverse problem (6) which is an optimization problem over $\mathfrak{S}$, it is reasonable to expect that an assumption relating the model $\mathfrak{S}$ and the data distribution $\pi$ might be required in order to control the excess risk. We now formalize this assumption. Let

$$\mathcal{S}^\kappa \triangleq \left\{ \frac{p-q}{\|\mathcal{A}(p-q)\|_\mathcal{F}} \;\middle|\; p, q \in \mathfrak{S}, \|\mathcal{A}(p-q)\|_\mathcal{F} > 0 \right\} \quad (12)$$

be the normalized secant of the model set $\mathfrak{S}$, which by definition is included in the unit sphere of $\mathcal{F}$. Given $t > 0$ define

$$\lambda_t \triangleq \sup_{\mu \in \mathcal{S}^\kappa} \inf \left\{ \lambda > 0 \;\middle|\; \mathcal{N}_{\mathcal{A}(\mu)}(\lambda) \leq t \right\}. \quad (13)$$

Note that, by construction, $\lambda_t$ is a decreasing function of $t$. Furthermore, for any $\mu \in \mathcal{S}^\kappa$ we have $\|\mathcal{A}(\mu)\|_\mathcal{F} = 1$ and thus $\mathcal{N}_{\mathcal{A}(\mu)}(\lambda) \leq 1/\lambda$, which implies $\inf \left\{ \lambda > 0 \;\middle|\; \mathcal{N}_{\mathcal{A}(\mu)}(\lambda) \leq t \right\} \leq \frac{1}{t}$ and $\lambda_t \leq \frac{1}{t}$. Moreover, since $\mathcal{N}_{\mathcal{A}(\mu)}(\lambda)$ is a continuous decreasing function of $\lambda$, it holds that

$$\mathcal{N}_{\mathcal{A}(\mu)}(\lambda) \leq t \qquad \forall \mu \in \mathcal{S}^\kappa, \lambda \geq \lambda_t \quad (14)$$

In order to better grasp the geometric meaning of this last equation, we use (12) and the definition of $\mathcal{N}_f$ to rewrite it as $\forall p, q \in \mathfrak{S}, \lambda \geq \lambda_t$

$$\|(\Sigma + \lambda I)^{-1/2} \mathcal{A}(p-q)\|_\mathcal{F} \leq \sqrt{t} \|\mathcal{A}(p-q)\|_\mathcal{F}. \quad (15)$$

The first term can be interpreted as a Mahalanobis distance between the mean embeddings $\mathcal{A}(p)$ and $\mathcal{A}(q)$ with respect to the operator $(\Sigma + \lambda_t I)^{1/2}$, which depends on both the feature map and the data distribution. The regularization term $\lambda$ is necessary here as the covariance operator $\Sigma$ might not be invertible. Eq. (15) states that Mahalanobis distance at $\lambda = \lambda_t$ between the mean embeddings $\mathcal{A}(p)$ and $\mathcal{A}(q)$ is bounded by above by the distance between $\mathcal{A}(p)$ and $\mathcal{A}(q)$ in $\mathcal{F}$. Notice that it always holds that $\lambda_t \leq 1/t$. We now assume a strict inequality, so that $(\Sigma + \lambda_t I)$ is closer to its limit $\Sigma$.

**Assumption 2.** *There exists $t^* > 0$ s.t. $3\lambda_{t^*} < 1/t^*$.*

The factor 3 is used to simplify the analysis and it is likely that a smaller constant could be used instead. Although the decay of the eigenvalues of the covariance operator $\Sigma$ can be characterized in some settings (e.g. geometric decay for a gaussian kernel and sub-gaussian data distribution (Widom 1963)), it is in general not possible to derive an expression of the associated eigenvectors outside of a few specific cases. For this reason, proving Assumption 2 for a model of interest is not straightforward. We will introduce in Proposition 4.1 a sufficient condition for it to hold, which will be slightly easier to interpret.

Finally, we will need an assumption to characterize to which extent the feature map $\Phi$ is compatible with the learning task to solve. For that, following Gribonval et al. (2021a, eq. 8) we define the following semi-norm associated to the loss function

$$\|\pi - \pi'\|_\mathcal{L} = \sup_{h \in H} |\mathcal{R}(\pi, h) - \mathcal{R}(\pi', h)|. \quad (16)$$

This definition naturally extends to finite signed measure via the Jordan decomposition. Note that other semi-norms can be used and might yield tighter bounds, but we stick with this definition for simplicity.

**Assumption 3.** *There exists $C_\mathcal{F} < \infty$ such that for any $p, q \in \mathfrak{S}, \|p - q\|_\mathcal{L} \leq C_\mathcal{F} \|\mathcal{A}(p-q)\|_\mathcal{F}$.*

This assumption does not involve our approximate feature map $\Phi_m$, and is already known to hold for mixtures of Diracs and mixtures of Gaussians with specific separation assumptions when using a Gaussian kernel (Gribonval et al. 2021b, Appendix D.2).

### 4.2 Main result

In order to state our main result, we introduce the following discrepancy between distributions

$$d_C(\pi_\mathfrak{S}, \pi) \triangleq 2\|\pi - \pi_\mathfrak{S}\|_\mathcal{L} + 4C\|\mathcal{A}(\pi_\mathfrak{S} - \pi)\|_\mathcal{F}. \quad (17)$$

Then we have the following result on the excess risk.

**Theorem 4.1 (Main result):** *Fix $\delta > 0$, let $\mathbf{X}$ be a set of $n$ samples drawn i.i.d. according to $\pi$, and $\tilde{\mathbf{X}}$ a set of $m$ landmarks drawn from $\mathbf{X}$ using either uniform or $(z, \lambda_0, \delta/2)$-ALS sampling. Fix a hypothesis space $H$, a model set $\mathfrak{S} \subset \mathcal{P}(\mathcal{X})$ and a feature map $\Phi : \mathcal{X} \to \mathcal{F}$ satisfying Assumptions 1 to 3. Define the estimator $\hat{h}$ by (7), using the feature map $\Phi_m$ derived from $\Phi$ and $\tilde{\mathbf{X}}$ as given in (8). Then with probability at least $1 - \delta$*

$$\text{ER}(\pi, \hat{h}) \leq \inf_{\pi_\mathfrak{S} \in \mathfrak{S}} d_C(\pi_\mathfrak{S}, \pi) + 4C\|\mathcal{A}(\pi - \pi_n)\|_\mathcal{F}, \quad (18)$$

*where $C \triangleq C_\mathcal{F}(1 - 3\lambda t)^{-1/2}$ and for any $t, \lambda$ satisfying*

$$\lambda \geq \lambda_t, \qquad (19a) \qquad 3\lambda t < 1, \qquad (19b)$$

*and provided that, depending on the setting:*

- *for uniform sampling*

$$m \geq \max(67, 5\mathcal{N}_\infty(\lambda)) \log \frac{4K^2}{\lambda \delta}. \qquad (20)$$

- *for ALS sampling*

$$m \geq \max(334, 78z^2 \mathcal{N}(\lambda)) \log \frac{16n}{\delta} \qquad (21a)$$

$$n \geq 1655K^2 + 233K^2 \log(4K^2/\delta) \qquad (21b)$$

$$\max\left(\lambda_0, \frac{19K^2}{n} \log\left(\frac{4n}{\delta}\right)\right) \leq \lambda \leq \|\Sigma\|_{\mathcal{L}(\mathcal{F})}. \qquad (21c)$$

Assumption 2 implies that there exists at least one pair $(t^*, \lambda_{t^*})$ satisfying Eqs. (19a) and (19b) (i.e. choosing $t = t^*$, $\lambda = \lambda_t = \lambda_{t^*}$). According to (20), the sketch size $m$ decreases for uniform sampling with $\lambda_t$ as it is of order $\mathcal{N}_\infty(\lambda_t) \log(1/\lambda_t)$, thus in order to find a good tradeoff between minimizing $m$ and the constant $C_\mathcal{F}/\sqrt{1-3\lambda t}$ in the bound one should choose $\lambda > \lambda_{t^*}$ and $t \neq t^*$. Note that, when using approximate leverage scores sampling, the sketch size grows with $\lambda$ only in $\mathcal{N}(\lambda)$, by opposition to $\mathcal{N}_\infty(\lambda) \log(1/\lambda)$ for uniform sampling. The first term in the bound (18) can be interpreted as a bias term, and we refer the reader to Gribonval et al. (2021a, Sec. 3.3) for a finer control of this term at least in the clustering setting. The second term can be controlled using a concentration inequality in $\mathcal{F}$ as we assumed the data to be sampled i.i.d. according to the data distribution $\pi$.

We now provide a sufficient condition such that Assumption 2 holds true. Recall that $\Sigma$ is a trace-class positive operator, hence by the Hilbert-Schmidt theorem there exists a base $(e_\ell)_\ell$ of $\mathcal{F}$ and a positive $\ell_1$-sequence $(\sigma_\ell)_\ell$ such that $\forall l, \Sigma e_\ell = \sigma_\ell e_\ell$. Without loss of generality, we assume $(\sigma_\ell)_\ell$ to be decreasing, and we have $\sigma_\ell \to 0$ as $l \to \infty$. We denote $\mathcal{I} = \{i \in \mathbb{N} \mid \sigma_i > 0\}$, which can be finite or only countable.

**Proposition 4.1:** *Assume there exist $s \in ]0, 1/2[$ and a constant $\gamma_s > 0$ such that*

$$\forall \mu \in \mathcal{S}^\kappa, \mathcal{A}(\mu) \in \Sigma^s \mathcal{F} \qquad (22a)$$

$$and \quad \sup_{\mu \in \mathcal{S}^\kappa} \sum_{\ell \in \mathcal{I}} \frac{\langle \mathcal{A}(\mu), e_\ell \rangle_\mathcal{F}^2}{\sigma_\ell^{2s}} \leq \gamma_s. \qquad (22b)$$

*Choose $t > (3^{1-2s} \gamma_s)^{1/(2s)}$ and define $\lambda = \left(\frac{\gamma_s}{t}\right)^{\frac{1}{1-2s}}$. Then Eqs. (19a) and (19b) are satisfied and the constant in Theorem 4.1 reads $C = C_\mathcal{F}(1 - 3\gamma_s^{\frac{1}{1-2s}} t^{\frac{2s}{1-2s}})^{-1/2}$.*

Note that (22b) is akin to the source conditions used in the literature on inverse problems (Engl et al. 2000).

### 4.3 Idea of the proof

Our goal is to control the excess risk (1) of the hypothesis $\hat{h}$ recovered from the sketch via a solution $\hat{\pi} \in \mathfrak{S}$ of the inverse problem (6). Following previous works on compressive learning with random features (Gribonval et al. 2021a,b), our strategy will be to show that the sketching operator satisfies a lower restricted isometry property (LRIP), i.e. that there exists a constant $C$ such that

$$\forall p, q \in \mathfrak{S}, \|p - q\|_\mathcal{L} \leq C \|\mathcal{A}_m(p-q)\|_2. \qquad (23)$$

We will see in Proposition 4.2 that (23) is a sufficient condition to control the excess risk of the recovered hypothesis. The motivation behind (23) comes from the compressive sensing literature. If we model the learning operation as $\hat{\pi} = \Delta(\mathcal{A}_m(\pi_n))$ for some "decoder" operator $\Delta : \mathcal{F} \to \mathfrak{S}$, and if we require $\Delta$ to be stable in the sense that for any probability distribution $\pi$ and noise $\mathbf{e} \in \mathbb{R}^m$, $\|\pi - \Delta(\mathcal{A}_m(\pi) + \mathbf{e})\|_\mathcal{L} \lesssim d(\pi, \mathfrak{S}) + \|\mathbf{e}\|_2$ for some measure $d(\cdot, \mathfrak{S})$ of the distance to the model set, then it can be shown that the LRIP (23) must hold with a finite constant; conversely if (23) holds then the moment-matching decoder $\Delta : s \mapsto \arg\min_{p \in \mathfrak{S}} \|\mathcal{A}_m(p) - s\|_\mathcal{F}$ can be shown to be stable (Bourrier et al. 2014).

We now characterize precisely how the excess risk can be controlled when the LRIP (23) holds. This result is adapted from Bourrier et al. (2014, Theorems 7&4).

**Proposition 4.2:** *Assume that the LRIP (23) holds with constant $C < \infty$. Then*

$$\mathrm{ER}(\pi, \hat{h}) \leq \inf_{\pi_\mathfrak{S} \in \mathfrak{S}} d_C(\pi_\mathfrak{S}, \pi) + 4C \|\mathcal{A}(\pi - \pi_n)\|_\mathcal{F}.$$

**Strategy to prove the LRIP** Our main result will be a direct consequence of Proposition 4.2, but it remains to prove that the LRIP (23) holds. One way to do so is to find constants $C_\mathcal{F}, C_\mathrm{a}$ such that the two following properties hold independently:

$$\forall p, q \in \mathfrak{S}, \|p-q\|_\mathcal{L} \leq C_\mathcal{F} \|\mathcal{A}(p-q)\|_\mathcal{F} \qquad (24)$$

$$\forall p, q \in \mathfrak{S}, \|\mathcal{A}(p-q)\|_\mathcal{F} \leq C_\mathrm{a} \|\mathcal{A}_m(p-q)\|_2. \qquad (25)$$

Here the first equation characterizes how the (pseudo)metric induced by the chosen kernel is compatible with the one induced by the loss. This equation is independent of the choice of the landmarks and already known to hold in our setting, which is why it is covered by Assumption 3. The second equation characterizes how the metric $\|\mathcal{A}(\cdot)\|_{\mathcal{F}}$ is approximated by its projected variant $\|P_m \mathcal{A}(\cdot)\|_{\mathcal{F}}$ for distributions in the model $\mathfrak{S}$, and we prove in Appendix C that it holds with high probability on the draw of the landmarks $\tilde{\mathbf{X}}$. Our proof differs from the the strategy followed in Gribonval et al. (2021b): while the latter proves a pointwise result (for $p, q \in \mathfrak{S}$ fixed) which is extended to the whole model set using covering arguments, we use instead a result from Rudi et al. (2015) which controls the interaction of the regularized covariance operator with the projection $P_m$. As a consequence we avoid the use of covering numbers, although some kind of uniformity is still induced by Assumption 2 as $\lambda_t$ is defined as a supremum over the secant set in (13).

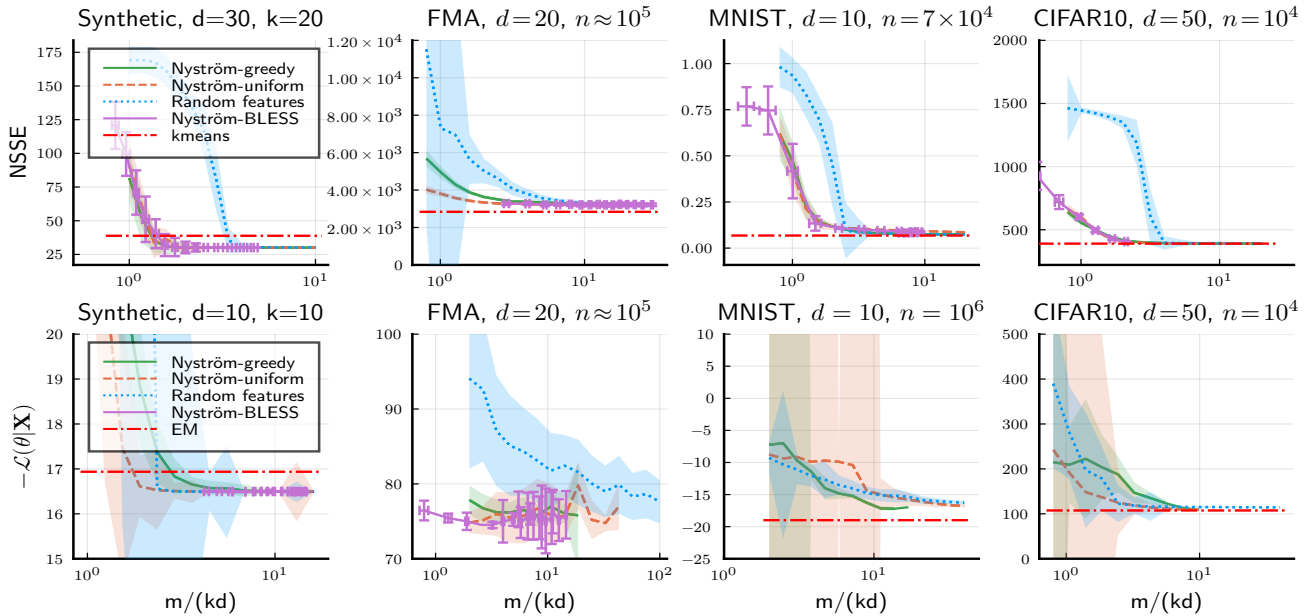

Figure 1: Median and standard deviation of the risk for 4 datasets for k-means clustering (top) and Gaussian modeling (bottom). For BLESS, the sketch size is randomized as well and each point corresponds to a single value of the parameter $\lambda$. See Appendix E for kernel parameters. We use $k = 10$ unless otherwise specified.

**Non-uniform LRIPs** Eqs. (24) and (25) are formulated as a uniform result for $p, q \in \mathfrak{S}$, but the excess risk can still be controlled using a weaker "non-uniform" result (Keriven et al. 2018, Theorem 2), i.e. showing that the corresponding inequalities hold for $p$ fixed and uniformly for $q \in \mathfrak{S}$. Although this might seem more natural here as the feature map is data-dependent, it did not allow us to derive better bounds.

## 5 EMPIRICAL RESULTS

We now compare the performance of compressive learning with Nyström and random features sketches.

The Nyström centers are sampled uniformly, according to ALS using BLESS, and according to the greedy iterative procedure described in Section 3.2. We perform both k-means clustering and Gaussian modeling experiments, and learning from the sketch is always performed using the CL-OMPR greedy heuristic (Keriven et al. 2017a,b). For this purpose we use the Julia CompressiveLearning package[1], to which we added the support of Nyström features. The source code to reproduce the experiments can be found online[2], and is also provided as supplementary material. We perform experiments on synthetic data drawn according to a Gaussian mixture, and on real datasets consisting in vectorial features extracted from the FMA (Defferrard et al. (2016), $d=20$ MFCC features), MNIST (LeCun et al. (1998), $d=10$) and CIFAR10 (Krizhevsky (2009), $d = 50$) datasets. We provide more details on data generation and features extraction in Appendix E. We use a Gaussian kernel $\kappa(\mathbf{x}, \mathbf{y}) = \exp(-\|\mathbf{x}-\mathbf{y}\|^2/(2\sigma^2))$ whose bandwidth $\sigma$ is manually chosen, and use $k = 10$ unless otherwise specified. In Figure 1 we report the risk as a function of the sketch size for both $k$-means clustering and Gaussian modeling. For clustering, the Nyström approximation consistently achieves lower error and standard deviation compared to random fea-tures, especially when using a small numbers of Nyström features. There does not seem to be a consistently better sampling strategy of the Nyström points. For Gaussian modeling, the Nyström approximation outperforms random features in terms of both median and standard deviation on the first two datasets. For MNIST, Nyström with uniform sampling has a very large standard deviation, but the greedy sampling strategy yields better results than random features (both in median and standard deviation). For CIFAR, Nyström seems to be on par with random features: uniform sampling yields a lower median error but a larger standard deviation. In Figure 2 we see the minus log-likelihood $-\mathcal{L}(\theta|\mathbf{X})$ as a function of both the sketch size $m$ and the kernel variance $\sigma$. It can be seen how the range of kernel variances yielding good results is wider for Nyström than for random features.

---
[1] https://gitlab.com/CompressiveLearning/CompressiveLearning.jl
[2] See https://gitlab.com/CompressiveLearning/mean-nystroem-embeddings-for-adaptive-compressive-learning-source-code-aistats-2022

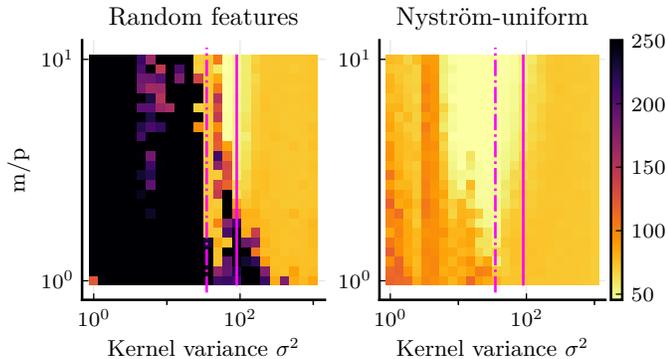

Figure 2: Median minus loglikelihood on a synthetic dataset ($k = 20$, $d = 30$) vs the sketch size (normalized by $p = 2kd$) and $\sigma^2$. Vertical lines correspond to the variances yielding the best results for each setting.

## 6 CONCLUSION AND PERSPECTIVES

We have introduced a new data-dependent sketch based on the Nyström method, and shown empirically that compressive k-means clustering and compressive Gaussian modeling can be performed using such sketches with much smaller sketch sizes than in previous works using random features. From a theoretical perspective, we provide a generic bound on the excess risk provided that the parametric model used to learn from the sketch is compatible with the data distribution and the feature map; we provide a sufficient condition for this to hold. It will be interesting in future works to prove that this condition holds in specific settings.

## Acknowledgments

The authors would like to thank Rémi Gribonval for helpful comments and discussions. Ernesto De Vito is a member of GNAMPA of JNdAM. Lorenzo Rosasco acknowledges the financial support of the European Research Council (grant SLING 819789), the AFOSR projects FA9550-18-1-7009, FA9550-17-1-0390 and BAA-AFRL-AFOSR-2016-0007 (European Office of Aerospace Research and Development), the EU H2020-MSCA-RISE project NoMADS - DLV-777826, and the Center for Brains, Minds and Machines (CBMM), funded by NSF STC award CCF-1231216.

# Supplementary Material: Mean Nyström Embeddings for Adaptive Compressive Learning

## A  COMPUTING THE GRADIENTS

### A.1  Mean sketch of an atomic Dirac measure (Gaussian kernel)

For an atomic dirac measure $P_\theta = \delta_\theta$, the computation is straightforward as

$$\mathcal{A}_m(P_\theta) = \Phi_m(\theta)$$

$$= \mathbf{K}_m^{-1/2} \begin{bmatrix} \kappa(\theta, \tilde{\mathbf{x}}_1) \\ \vdots \\ \kappa(\theta, \tilde{\mathbf{x}}_m) \end{bmatrix}$$

**Jacobian**  Denoting $F(\theta, \tilde{\mathbf{x}}) = \mathbf{E}_{x \sim P_\theta}\kappa(x, \tilde{\mathbf{x}}) = \kappa(\theta, \tilde{\mathbf{x}}) = \exp\left(-\frac{\|\theta - \tilde{\mathbf{x}}\|^2}{2\sigma^2}\right)$, we have

$$\frac{\partial F(\theta, \tilde{\mathbf{x}})}{\partial \theta} = -\frac{1}{\sigma^2}(\theta - \tilde{\mathbf{x}})F(\theta, \tilde{x}) \tag{26}$$

Denoting $f : \theta \to [F(\theta, \tilde{\mathbf{x}}_1), ..., F(\theta, \tilde{\mathbf{x}}_m)]^T$ and $z = \mathbf{K}_m^{-1/2}y$ for $y \in \mathbb{R}^m$, we have:

$$(J_{\mathcal{A}_m}(\theta))^T y = \left[\frac{\partial F(\theta, \tilde{x}_1)}{\partial \theta}, ..., \frac{\partial F(\theta, \tilde{x}_m)}{\partial \theta}\right] \mathbf{K}_m^{-1/2} y \tag{27}$$

$$= -\frac{1}{\sigma^2}[(\theta - \tilde{\mathbf{x}}_i)F(\theta, \tilde{\mathbf{x}}_i)]_{1 \leq i \leq m} z \tag{28}$$

$$= -\frac{1}{\sigma^2}\left[f(\theta)^T z \theta - \tilde{\mathbf{X}}(z \odot f(\theta))\right] \in \mathbb{R}^d \tag{29}$$

### A.2  Mean sketch of an atomic Gaussian distribution with diagonal covariance (Gaussian kernel)

Although we considered in Example 2 Gaussian mixtures with a fixed known covariance matrix, we derive here more general rules for Gaussian mixtures with (learnable) diagonal covariance matrices.

Let $F(\theta, \tilde{\mathbf{x}}) = k(P_\theta, \tilde{\mathbf{x}}) \triangleq \mathbf{E}_{\mathbf{x} \sim P_\theta} k(\mathbf{x}, \tilde{\mathbf{x}})$. Let $|\cdot|$ denote the determinant. We have:

$$F(\theta = (\boldsymbol{\mu}, \boldsymbol{\Gamma}), \tilde{\mathbf{x}}) = E_{\mathbf{x} \sim \mathcal{N}(\boldsymbol{\mu}, \boldsymbol{\Gamma})} k(\mathbf{x}, \tilde{\mathbf{x}})$$

$$= \int N(\mathbf{x}; \boldsymbol{\mu}, \boldsymbol{\Gamma})|2\pi\sigma^2 I|^{1/2} \mathcal{N}(\mathbf{x}; \tilde{\mathbf{x}}, \sigma^2 I) d\mathbf{x}$$

$$\stackrel{\text{Petersen et al. 2012, eq. (371)}}{=} |2\pi\sigma^2 I|^{\frac{1}{2}} \mathcal{N}(\tilde{\mathbf{x}}; \boldsymbol{\mu}, \boldsymbol{\Gamma} + \sigma^2 I) \left(\int N(\mathbf{m}_c, \boldsymbol{\Gamma}_c) dx\right)$$

$$= |2\pi\sigma^2 I|^{\frac{1}{2}} \mathcal{N}(y; \tilde{\mathbf{x}}, \boldsymbol{\Gamma} + \sigma^2 I)$$

$$= \sigma^d \left(\prod_{i=1}^d \sigma_i^2 + \sigma^2\right)^{-1/2} \exp(-\tfrac{1}{2}(\tilde{\mathbf{x}} - \boldsymbol{\mu})^T(\boldsymbol{\Gamma} + \sigma^2 I)^{-1}(\tilde{\mathbf{x}} - \boldsymbol{\mu}))$$

As a consequence:

$$\mathcal{A}_m(P_{\theta=(\boldsymbol{\mu}, \boldsymbol{\Gamma})}) = \frac{\sigma^d}{\left(\prod_{i=1}^d \sigma_i^2 + \sigma^2\right)^{1/2}} K^{-1/2} \begin{bmatrix} \exp(-\tfrac{1}{2}(\tilde{\mathbf{x}}_1 - \boldsymbol{\mu})^T(\boldsymbol{\Gamma} + \sigma^2 I)^{-1}(\tilde{\mathbf{x}}_1 - \boldsymbol{\mu})) \\ \vdots \\ \exp(-\tfrac{1}{2}(\tilde{\mathbf{x}}_m - \boldsymbol{\mu})^T(\boldsymbol{\Gamma} + \sigma^2 I)^{-1}(\tilde{\mathbf{x}}_m - \boldsymbol{\mu})) \end{bmatrix} \tag{30}$$

Antoine Chatalic, Luigi Carratino, Ernesto De Vito, Lorenzo Rosasco

**Jacobian** In the following, we use the notation $J_{\mathcal{A}_m} \triangleq J_{\theta \to \mathcal{A}_m(P_\theta)}$. Note that $\mathcal{A}_m(P_\theta) = K^{-1/2} f(\theta)$ with $f : \theta \to [F(\theta, \tilde{\mathbf{x}}_1), ..., F(\theta, \tilde{\mathbf{x}}_m)]^T$, hence $J_{\mathcal{A}_m}(\theta) = K^{-1/2} J_f(\theta)$.

The gradient of $F$ with respect to $\boldsymbol{\mu}$ and $\boldsymbol{\Gamma}$ (considered as a vector) is then:

$$\frac{\partial F(\theta, \tilde{\mathbf{x}})}{\partial \boldsymbol{\mu}} = -F(\theta, \tilde{\mathbf{x}})(\boldsymbol{\Gamma} + \sigma^2 I)^{-1}(\boldsymbol{\mu} - \tilde{\mathbf{x}})$$

$$\frac{\partial F(\theta, \tilde{\mathbf{x}})}{\partial \boldsymbol{\Gamma}} = \tfrac{1}{2} F(\theta, \tilde{\mathbf{x}})(\boldsymbol{\Gamma} + \sigma^2 I)^{-1}((\boldsymbol{\Gamma} + \sigma^2 I)^{-1}(\boldsymbol{\mu} - \tilde{\mathbf{x}})^{\odot 2} - 1),$$

where we use a $\odot$ to denote pointwise multiplication.

Denote $\boldsymbol{\Gamma}_{v,\sigma}^{-1} \triangleq \text{vec}((\boldsymbol{\Gamma} + \sigma^2 I)^{-1})$, and decompose the jacobian of $f$ in $J_f = [J_f^{\boldsymbol{\mu}}, J_f^{\boldsymbol{\Gamma}}] \in \mathbb{R}^{m \times 2d}$. For efficient computation, we need an expression for any vector $y$ of:

$$(J_f^{\boldsymbol{\mu}})^T y = \left[ \frac{\partial F(\theta, \tilde{\mathbf{x}}_1)}{\partial \boldsymbol{\mu}}, ..., \frac{\partial F(\theta, \tilde{\mathbf{x}}_m)}{\partial \boldsymbol{\mu}} \right] y \tag{31}$$

$$= -\left[ F(\theta, \tilde{\mathbf{x}}_1) \boldsymbol{\Gamma}_{v,\sigma}^{-1} \odot (\boldsymbol{\mu} - \tilde{\mathbf{x}}_1), ..., F(\theta, \tilde{\mathbf{x}}_1) \boldsymbol{\Gamma}_{v,\sigma}^{-1} \odot (\boldsymbol{\mu} - \tilde{\mathbf{x}}_1) \right] y \tag{32}$$

$$= \boldsymbol{\Gamma}_{v,\sigma}^{-1} \odot ([F(\theta, \tilde{\mathbf{x}}_1) \tilde{\mathbf{x}}_m, ..., F(\theta, \tilde{\mathbf{x}}_1) \tilde{\mathbf{x}}_m] y) - \boldsymbol{\Gamma}_{v,\sigma}^{-1} \odot \boldsymbol{\mu}(\sum_i y_i F(\theta, \tilde{\mathbf{x}}_i)) \tag{33}$$

$$= \boldsymbol{\Gamma}_{v,\sigma}^{-1} \odot \left( \tilde{\mathbf{X}}(y \odot f(\theta)) - \boldsymbol{\mu}(\sum_i y_i F(\theta, \tilde{\mathbf{x}}_i)) \right) \tag{34}$$

And with respect to $\boldsymbol{\Gamma}$, we have:

$$(J_f^{\boldsymbol{\mu}})^T y = \left[ \frac{\partial F(\theta, \tilde{\mathbf{x}}_1)}{\partial \boldsymbol{\Gamma}}, ..., \frac{\partial F(\theta, \tilde{\mathbf{x}}_m)}{\partial \boldsymbol{\Gamma}} \right] y \tag{35}$$

$$= \tfrac{1}{2} \boldsymbol{\Gamma}_{v,\sigma}^{-1} \odot \left[ (-1 + \boldsymbol{\Gamma}_{v,\sigma}^{-1} \odot \boldsymbol{\mu}^{\odot 2}) + (\boldsymbol{\Gamma}_{v,\sigma}^{-1} \odot \tilde{\mathbf{x}}_i^{\odot 2} - 2 \boldsymbol{\Gamma}_{v,\sigma}^{-1} \odot \boldsymbol{\mu} \odot \tilde{\mathbf{x}}_i) \right]_{1 \le i \le m} (y \odot f(\theta)) \tag{36}$$

$$= \tfrac{1}{2} \boldsymbol{\Gamma}_{v,\sigma}^{-1} \odot \left( (-1 + \boldsymbol{\Gamma}_{v,\sigma}^{-1} \odot \boldsymbol{\mu}^{\odot 2})(\sum_i (y \odot f(\theta))_i) + \boldsymbol{\Gamma}_{v,\sigma}^{-1} \odot (\tilde{X}^{\odot 2} - 2\boldsymbol{\mu} \odot \tilde{X})(y \odot f(\theta)) \right) \tag{37}$$

## B  SKETCHING OPERATOR

Fix a locally compact second countable topological space $\mathcal{X}$ endowed with its Borel-$\sigma$ algebra $\mathcal{B}$. We set

a) $\mathcal{C}_0(\mathcal{X})$ be the Banach space of continuous functions $f : \mathcal{X} \to \mathbb{R}$ going to zero at infinity, endowed with the sup norm $\|f\|_\infty$;

b) $\mathcal{L}_b(\mathcal{X})$ be the Banach space of bounded Borel measurable functions $f : \mathcal{X} \to \mathbb{R}$ endowed with the sup norm $\|f\|_\infty$;

c) $\mathcal{M}(\mathcal{X})$ be the Banach space of finite signed measures on $\mathcal{X}$ endowed the total variation norm $\|\mu\|_{\mathrm{TV}}$;

d) $\mathcal{M}(\mathcal{X})_+ \subset \mathcal{M}(\mathcal{X})$ be the cone of positive measures;

e) $\mathcal{P}(\mathcal{X}) \subset \mathcal{M}(\mathcal{X})_+$ be the convex subset $\mathcal{M}(\mathcal{X})$ of probability measures on $\mathcal{X}$.

We recall the following standard facts.

i) given a signed measure $\mu \in \mathcal{M}(\mathcal{X})$, there exists a unique positive measure $|\mu|$, called the absolute value of $\mu$ and a (almost unique) function $\Delta_\mu : \mathcal{X} \to \{\pm 1\}$, called the Radon-Nikodym derivative, such that

$$\mu(E) = \int_E \Delta_\mu(x) \, d|\mu|(x) \qquad E \in \mathcal{B}$$

and $\|\mu\|_{\mathrm{TV}} = |\mu|(\mathcal{X})$;

ii) given a function $\varphi$, which is integrable with respect to $|\mu|$, the integral of $f$ with respect to $\mu$ is given by

$$\int_\mathcal{X} \varphi(x) \, d\mu(x) = \int_\mathcal{X} \varphi(x) \Delta_\mu(x) \, d|\mu|(x), \tag{38}$$

which is equivalent to the definition in terms of Hahn decomposition;

iii) $\mathcal{M}(\mathcal{X})$ can be identified, as a Banach space, with the dual $\mathcal{C}_0(\mathcal{X})^*$ of $\mathcal{C}_0(\mathcal{X})$ by the duality pairing

$$\langle \mu, \varphi \rangle_{\mathcal{C}_0(\mathcal{X})} = \int_{\mathcal{X}} \varphi(x) \Delta_\mu(x) \, d|\mu|(x) \qquad \mu \in \mathcal{M}(\mathcal{X}), \ \varphi \in \mathcal{C}_0(\mathcal{X}); \tag{39}$$

iv) for all $\mu \in \mathcal{M}(\mathcal{X})$

$$\|\mu\|_{\mathrm{TV}} = \sup_{\varphi \in \mathcal{C}_0(\mathcal{X})} \int_{\mathcal{X}} \varphi(x) \, d\mu(x) = \sup_{\varphi \in \mathcal{L}_b(\mathcal{X})} \int_{\mathcal{X}} \varphi(x) \, d\mu(x),$$

so that $M(\mathcal{X})$ is a closed subspace of $\mathcal{L}_b(\mathcal{X})^*$, where the duality pairing

$$\langle \mu, \varphi \rangle_{\mathcal{L}_b(\mathcal{X})} = \int_{\mathcal{X}} \varphi(x) \Delta_\mu(x) \, d|\mu|(x),$$

where $\mu \in \mathcal{M}(\mathcal{X})$ and $\varphi \in \mathcal{L}_b(\mathcal{X})$.

Take a separable Hilbert space $\mathcal{F}$ and a bounded measurable map $\Phi : \mathcal{X} \to \mathcal{F}$ and define the bounded operator

$$S_\Phi : \mathcal{F} \to \mathcal{L}_b(\mathcal{X}) \qquad (S_\Phi f)(x) = \langle f, \Phi(x) \rangle_\mathcal{F},$$

with operator norm $\|S_\Phi\| = \sup_{x \in \mathcal{X}} \|\Phi(x)\|_\mathcal{F}$. The adjoint $S_\Phi^*$ is a bounded operator from the dual $\mathcal{L}_b(\mathcal{X})^*$ into $\mathcal{F}$. Hence, by item iv) above, its restriction to $\mathcal{M}(\mathcal{X})$

$$\mathcal{A}_\Phi : \mathcal{M}(\mathcal{X}) \to \mathcal{F} \qquad \mathcal{A}_\Phi \mu = S_\Phi^* \mu$$

is continuous too, with $\|\mathcal{A}_\Phi\| = \sup_{x \in \mathcal{X}} \|\Phi(x)\|_\mathcal{F}$. Furthermore, it holds that

$$\langle \mathcal{A}_\Phi \mu, f \rangle_\mathcal{F} = \mu(S_\Phi f) = \int_{\mathcal{X}} \langle f, \Phi(x) \rangle_\mathcal{F} \, d\mu(x) = \int_{\mathcal{X}} \langle \Phi(x), f \rangle_\mathcal{F} \Delta_\mu(x) \, d|\mu|(x) \qquad f \in \mathcal{F}. \tag{40}$$

Since the maps $\Phi$ and $\Delta_\mu$ are bounded and measurable, if follows that

$$\mathcal{A}_\Phi \mu = \int_{\mathcal{X}} \Phi(x) \Delta_\mu(x) d|\mu|(x), \tag{41}$$

where the integral is the vector valued Bochner integral. Note that if $\pi$ is a probability measure the above equation reads as

$$\mathcal{A}_\Phi \pi = \int_{\mathcal{X}} \Phi(x) d\pi(x) = \mathbf{E}_{\mathbf{x} \sim \pi} \Phi(x),$$

which is usually called the kernel mean embedding. Furthermore, if $\mu \in \mathcal{M}(\mathcal{X})_+$ is a positive measure, there exists a natural continuous embedding[3]

$$\iota_\mu : \mathcal{L}_b(\mathcal{X}) \hookrightarrow L^2(\mathcal{X}, \mu) \qquad (\iota_\mu \varphi)(x) = \varphi(x) \qquad \text{for } \mu\text{-almost all } x \in \mathcal{X},$$

whose adjoint[4] $\iota^*$ takes value in $\mathcal{M}(\mathcal{X})$ and it is given by

$$\iota_\mu^* : L^2(\mathcal{X}, \mu) \hookrightarrow \mathcal{M}(\mathcal{X}) \qquad \iota_\mu^* F = F \cdot \mu,$$

where $F \cdot \mu$ is the signed measure having density $F$ with respect to $\mu$. Furthermore, the operator $S_{\Phi, \mu} = \iota_\mu S_\Phi$

$$S_{\Phi,\mu} : \mathcal{F} \to L^2(\mathcal{X}, \mu) \qquad S_{\Phi,\mu} f(x) = \langle f, \Phi(x) \rangle_\mathcal{F} \qquad \text{for } \mu\text{-almost all } x \in \mathcal{X}$$

is the *restriction operator* and its adjoint $S_{\Phi,\mu}^* = S_\Phi^* \iota_\mu^*$ is the *extension operator*

$$S_{\Phi,\mu}^* : L^2(\mathcal{X}, \mu) \to \mathcal{F} \qquad S_{\Phi,\mu}^* F = \int_{\mathcal{X}} \Phi(x) F(x) \, d\mu(x).$$

---

[3] Since the elements of $L^2(\mathcal{X}, \mu)$ are equivalence classes of function, $\iota$ is not injective, so that the notation $\hookrightarrow$ is a little bit misleading.

[4] Since $\iota$ has dense range, $\iota^*$ is injective, so that $\iota^*$ is a true embedding.

It is known that $S_{\Phi,\mu}^* S_{\Phi,\mu} : \mathcal{F} \to \mathcal{F}$ is given by

$$S_{\Phi,\mu}^* S_{\Phi,\mu} = \int_{\mathcal{X}} \Phi(x) \otimes \Phi(x), d\mu(x),$$

where the integral is a vector valued Bochner integral taking value in the Hilbert space of Hilbert-Schmidt operators and it is a positive trace class operator (De Vito et al. 2014, Proposition 14). Furthermore, $S_{\Phi,\mu} S_{\Phi,\mu}^*$ is the integral operator on $L^2(\mathcal{X}, \mu)$ with kernel

$$\kappa(x, x') = \langle \Phi(x), \Phi(x') \rangle_{\mathcal{F}}.$$

In particular, if $\pi$ is a probability measure

$$S_{\Phi,\pi}^* S_{\Phi,\pi} = \mathbf{E}_{x \sim \pi} [\Phi(x) \otimes \Phi(x)]$$

is the (non-centered) covariance operator. Note that if $\mu$ is any finite signed measure in $\mathcal{M}(\mathbf{X})$, then

$$\mathcal{A}_\Phi \mu = S_{\Phi,|\mu|}^* \Delta_\mu,$$

where clearly $\Delta_\mu \in L^2(\mathcal{X}, |\mu|)$ .

## C PROOFS

### C.1 Proofs of Section 3

**Proof of Lemma 3.1:** *We introduce the operator*

$$\mathbf{\Phi}_{\tilde{\mathbf{X}}} : \mathbb{R}^m \to \mathcal{F}, \mathbf{a} \mapsto \sum_{i=1}^m a_i \Phi(\tilde{\mathbf{x}}_i), \quad \text{with adjoint} \quad \mathbf{\Phi}_{\tilde{\mathbf{X}}}^* : f \mapsto [\langle f, \Phi(\tilde{\mathbf{x}}_1) \rangle_{\mathcal{F}}, ..., \langle f, \Phi(\tilde{\mathbf{x}}_m) \rangle_{\mathcal{F}}]^T. \tag{42}$$

*It is easy to check that*

$$\operatorname{Im}(\mathbf{\Phi}_{\tilde{\mathbf{X}}}) = \mathcal{F}_m \qquad \ker(\mathbf{\Phi}_{\tilde{\mathbf{X}}}) = \ker(\mathbf{K}_m) \qquad \mathbf{\Phi}_{\tilde{\mathbf{X}}}^* \mathbf{\Phi}_{\tilde{\mathbf{X}}} = \mathbf{K}_m.$$

*The polar decomposition of $\mathbf{\Phi}_{\tilde{\mathbf{X}}}$ reads*

$$\mathbf{\Phi}_{\tilde{\mathbf{X}}} = U \mathbf{K}_m^{1/2}$$

*where $U : \mathbb{R}^d \to \mathcal{F}$ satisfies the equations*

$$U^* U \mathbf{c} = \mathbf{c} \qquad \forall \mathbf{c} \in \ker(\mathbf{K}_m)^\perp \tag{43a}$$
$$UU^* f = f \qquad \forall f \in \mathcal{F}_m \tag{43b}$$
$$U\mathbf{c} = 0 \qquad \forall \mathbf{c} \in \ker(\mathbf{K}_m) \tag{43c}$$
$$U^* f = 0 \qquad \forall f \in \mathcal{F}_m^\perp, \tag{43d}$$

*i.e. it is a partial isometry from $\ker(\mathbf{K}_m)^\perp$ onto $\mathcal{F}_m$. By definition of $\Phi_m$, for all $\mathbf{x} \in \mathcal{X}$*

$$\Phi_m(\mathbf{x}) = \mathbf{K}_m^{-1/2} \mathbf{\Phi}_{\tilde{\mathbf{X}}}^* \Phi(\mathbf{x}) = \mathbf{K}_m^{-1/2} (U \mathbf{K}_m^{1/2})^* \Phi(\mathbf{x}) = \mathbf{K}_m^{-1/2} \mathbf{K}_m^{1/2} U^* \Phi(\mathbf{x}) = U^* \Phi(\mathbf{x}) \tag{44}$$

*where the last equality is due to the fact that $\mathbf{K}_m^{-1/2} \mathbf{K}_m^{1/2} \mathbf{c} = \mathbf{c}$ for every $\mathbf{c} \in \ker(\mathbf{K}_m)^\perp = \operatorname{Im}(U)$ by (43a). We have*

$$U \Phi_m(\mathbf{x}) = UU^* \Phi(\mathbf{x}) = P_m \Phi(\mathbf{x})$$

*by (43b), which gives (11a). Eqs. (4) and (11a) together give (11b). Finally, for any $\mathbf{x}, \mathbf{y} \in \mathcal{X}$ we have by (44)*

$$\langle \Phi_m(\mathbf{x}), \Phi_m(\mathbf{y}) \rangle = \langle U^* \Phi(\mathbf{x}), U^* \Phi(\mathbf{y}) \rangle = \langle \Phi(\mathbf{x}), UU^* \Phi(\mathbf{y}) \rangle_{\mathcal{F}} = \langle \Phi(\mathbf{x}), P_m \Phi(\mathbf{y}) \rangle_{\mathcal{F}}$$

*which yields (11c) as $P_m$ is a projection.*

### C.2 Proof of the main result when sampling uniformly the landmarks

**Proof of Proposition 4.2:** *First, note that the excess risk can be bounded using $\|\cdot\|_{\mathcal{L}}$ as follows (remember $\hat{h} \in \arg\min_h \mathcal{R}(\hat{\pi}, h)$):*

$$\mathrm{ER}(\pi, \hat{h}) \triangleq \mathcal{R}(\pi, \hat{h}) - \mathcal{R}(\pi, h^*) \tag{45}$$

$$= (\mathcal{R}(\pi, \hat{h}) - \mathcal{R}(\hat{\pi}, \hat{h})) + (\mathcal{R}(\hat{\pi}, \hat{h}) - \mathcal{R}(\pi, h^*)) \tag{46}$$

$$\stackrel{(i)}{\leq} (\mathcal{R}(\pi, \hat{h}) - \mathcal{R}(\hat{\pi}, \hat{h})) + (\mathcal{R}(\hat{\pi}, h^*) - \mathcal{R}(\pi, h^*)) \tag{47}$$

$$\leq 2 \sup_h |\mathcal{R}(\pi, h) - \mathcal{R}(\hat{\pi}, h)| \tag{48}$$

$$= 2\|\pi - \hat{\pi}\|_{\mathcal{L}} \tag{49}$$

*where $(i)$ follows from the definition of $\hat{h}$. Denoting $y = \mathcal{A}_m(\pi_n)$, we have:*

$$\|\pi - \hat{\pi}\|_{\mathcal{L}} \leq \|\pi - \pi_{\mathfrak{S}}\|_{\mathcal{L}} + \|\pi_{\mathfrak{S}} - \hat{\pi}\|_{\mathcal{L}} \tag{50}$$

$$\leq \|\pi - \pi_{\mathfrak{S}}\|_{\mathcal{L}} + C\|\mathcal{A}_m(\pi_{\mathfrak{S}} - \hat{\pi})\|_2 \tag{51}$$

$$\leq \|\pi - \pi_{\mathfrak{S}}\|_{\mathcal{L}} + C(\|\mathcal{A}_m(\pi_{\mathfrak{S}}) - y\|_2 + \|y - \mathcal{A}_m(\hat{\pi})\|_2) \tag{52}$$

$$\stackrel{(ii)}{\leq} \|\pi - \pi_{\mathfrak{S}}\|_{\mathcal{L}} + 2C\|\mathcal{A}_m(\pi_{\mathfrak{S}}) - y\|_2 \tag{53}$$

$$\leq \|\pi - \pi_{\mathfrak{S}}\|_{\mathcal{L}} + 2C(\|\mathcal{A}_m(\pi_{\mathfrak{S}}) - \mathcal{A}_m(\pi)\|_2 + \|\mathcal{A}_m(\pi) - y\|_2) \tag{54}$$

$$= [\|\pi - \pi_{\mathfrak{S}}\|_{\mathcal{L}} + 2C\|\mathcal{A}_m(\pi_{\mathfrak{S}}) - \mathcal{A}_m(\pi)\|_2] + 2C\|\mathcal{A}_m(\pi) - y\|_2 \tag{55}$$

*Where $(ii)$ follows from the definition of the decoder $\Delta$. By Lemma 3.1, we have*

$$\|\mathcal{A}_m(\pi_{\mathfrak{S}}) - \mathcal{A}_m(\pi)\|_2 = \|P_m \mathcal{A}(\pi_{\mathfrak{S}} - \pi)\|_{\mathcal{F}} \leq \|\mathcal{A}(\pi_{\mathfrak{S}} - \pi)\|_{\mathcal{F}} = \|\mathcal{A}(\pi_{\mathfrak{S}} - \pi)\|_{\mathcal{F}}$$

*as $P_m$ is a projection, and for the same reason $\|\mathcal{A}_m(\pi - \pi_n)\|_2 \leq \|\mathcal{A}(\pi - \pi_n)\|_{\mathcal{F}}$.*

In order to prove Theorem 4.1, we need the following result from Rudi et al. (2015)

**Lemma C.1 (Rudi et al. (2015, Lemma 6)):** *Under Assumption 1, when the set of $m$ landmarks is drawn uniformly from all partitions of cardinality $m$, for any $\lambda > 0$ we have*

$$\|P_m^\perp (\Sigma + \lambda I)^{1/2}\|_{\mathcal{L}(\mathcal{F})}^2 \leq 3\lambda$$

*with probability $1 - \delta$ provided*

$$m \geq \max(67, 5\mathcal{N}_\infty(\lambda)) \log \frac{4K^2}{\lambda \delta}.$$

Note that although the lemma is formulated for sampling without replacement, yet the proof seems to rely on a concentration result for i.i.d. sampling. We thus only formulate our result for i.i.d. sampling by precaution, but in practice a similar result should hold when sampling without replacement using an adapted concentration inequality, and this should only help to improve the constants.

**Proof of Theorem 4.1:** *To avoid ambiguity, we prove here the result for uniform sampling only and formulate just below a separated Theorem C.1 for the ALS setting. The claim is a direct consequence of Proposition 4.2 provided we prove that (23) holds with high probability. Note that when Eqs. (24) and (25) both hold with respective constants $C_{\mathcal{F}}$ and $C_a$, Eq. (23) holds with constant $C = C_{\mathcal{F}} C_a$. As Eq. (24) already holds by (3), we only need to prove (25). Fix $\delta$, fix $t$, $\lambda$ and $m$ satisfying Eqs. (19a), (19b) and (20) and define $\varepsilon = \sqrt{3\lambda t}$. Observe that with probability $1 - \delta$ on the draw of the Nyström landmarks*

$$\forall \mu \in \mathcal{S}^\kappa, \|P_m^\perp \mathcal{A}(\mu)\|_{\mathcal{F}} \leq \|P_m^\perp (\Sigma + \lambda I)^{1/2}\|_{\mathcal{L}(\mathcal{F})} \|(\Sigma + \lambda I)^{-1/2} \mathcal{A}(\mu)\|_{\mathcal{F}} \tag{56}$$

$$\stackrel{(i)}{\leq} \sqrt{3\lambda} \|(\Sigma + \lambda I)^{-1/2} \mathcal{A}(\mu)\|_{\mathcal{F}} \tag{57}$$

$$\stackrel{(ii)}{\leq} \sqrt{3\lambda} \sqrt{t} = \varepsilon \tag{58}$$

*where (i) is a consequence of Lemma C.1 taking into account that m satisfies* (20), *and (ii) comes from Eq.* (15) *taking into account that* $\lambda \geq \lambda_t$. *By the Pythagorean theorem and* (11b) *we get*

$$\forall \mu \in \mathcal{S}^\kappa, \|\mathcal{A}(\mu)\|_{\mathcal{F}}^2 = 1 = \|P_m \mathcal{A}(\mu)\|_{\mathcal{F}}^2 + \|P_m^\perp \mathcal{A}(\mu)\|_{\mathcal{F}}^2 = \|\mathcal{A}_m(\mu)\|_2^2 + \|P_m^\perp \mathcal{A}(\mu)\|_{\mathcal{F}}^2 \tag{59}$$

*so that using* (58) *we obtain* $1 - \varepsilon^2 \leq \|\mathcal{A}_m(\mu)\|_2^2$, *i.e.* (25) *holds with constant* $C_a = (1-\varepsilon^2)^{-1/2}$. *Note that* $\epsilon < 1$ *by* (19b).

### C.3 Faster rates with leverage scores sampling

We now explain how our result can be adapted when sampling the landmarks according to approximate leverage scores. For this we rely on the Lemma 7 from Rudi et al. (2015) (where a square seems to be missing in the lemma's statement, and we again omit the decay assumption on the effective dimension which is not used in the proof). We first recall this lemma for clarity.

**Lemma C.2 (ALS NyStröm approximation (Rudi et al. 2015, Lemma 7)):** *Let* $\lambda > 0$ *and* $\delta > 0$. *Let* $(\hat{l}_i(t))_{1 \leq i \leq n}$ *be a collection of* $(z, \lambda_0, \delta)$-*approximate leverage scores as defined in Section 3.2 for some* $z \geq 1$ *and* $\lambda_0 > 0$. *Let* $p_\lambda$ *be a probability distribution on the set of indexes* $\{1, ..., n\}$ *defined as* $p_\lambda(i) \triangleq \hat{l}_i(\lambda)/(\sum_{i=1}^n \hat{l}_i(\lambda))$. *Let* $\{i_1, ..., i_m\}$ *be a collection of indices sampled independently with replacement from* $p_\lambda$, *and* $P_m$ *the orthogonal projection on* $\mathcal{F}_m = \text{span}\{\Phi(\mathbf{x}_{i_1}), ..., \Phi(\mathbf{x}_{i_m})\}$. *Then we have with probability* $1 - 2\delta$

$$\|P_m^\perp (\Sigma + \lambda I)^{1/2}\|_{\mathcal{L}(\mathcal{F})}^2 \leq 3\lambda$$

*provided that:*
- *assumption 1 hold;*
- $n \geq 1655K^2 + 233K^2 \log(2K^2/\delta)$;
- $\max\left(\lambda_0, \frac{19K^2}{n} \log(\frac{2n}{\delta})\right) \leq \lambda \leq \|\Sigma\|_{\mathcal{L}(\mathcal{F})}$;
- $m \geq \max(334, 78z^2 \mathcal{N}(\lambda)) \log \frac{8n}{\delta}$.

**Theorem C.1:** *Let* $\mathbf{X}$ *be a set of* $n$ *samples drawn i.i.d. according to* $\pi$. *Let* $(\hat{l}_i(t))_{1 \leq i \leq n}$ *be a collection of* $(z, \lambda_0, \delta)$-*approximate leverage scores (cf. Section 3.2) for some* $z \geq 1$ *and* $\lambda_0 > 0$. *Let* $\lambda > 0$, *and* $p_\lambda$ *be a probability distribution on the set of indexes* $\{1, ..., n\}$ *defined as* $p_\lambda(i) \triangleq \hat{l}_i(\lambda)/(\sum_{i=1}^n \hat{l}_i(\lambda))$. *Let* $\{i_1, ..., i_m\}$ *be a collection of indices sampled independently with replacement from* $p_\lambda$, *and* $\tilde{\mathbf{X}}$ *the corresponding set of landmarks (without duplicates).*

*Fix a hypothesis space* H, *a model set* $\mathfrak{S} \subset \mathcal{P}(\mathcal{X})$ *and a feature map* $\Phi : \mathcal{X} \to \mathcal{F}$ *satisfying Assumptions 1 to 3. Define the estimator* $\hat{h}$ *by* (7), *where we implicitly use the feature map* $\Phi_m$ *derived from* $\Phi$ *and* $\tilde{\mathbf{X}}$ *as given in* (8). *Fix* $\delta > 0$, *with probability at least* $1 - 2\delta$

$$\text{ER}(\pi, \hat{h}) \leq \inf_{\pi_\mathfrak{S} \in \mathfrak{S}} d_{C'}(\pi_\mathfrak{S}, \pi) + 4C' \|\mathcal{A}(\pi - \pi_n)\|_{\mathcal{F}} \quad \text{where} \quad C' \triangleq \frac{C_\mathcal{F}}{\sqrt{1 - 3\lambda t}} \tag{60}$$

*provided that*

$$\lambda \geq \lambda_t \tag{61a}$$
$$3\lambda t < 1 \tag{61b}$$
$$m \geq \max(334, 78z^2 \mathcal{N}(\lambda)) \log \frac{8n}{\delta} \tag{61c}$$
$$n \geq 1655K^2 + 233K^2 \log(2K^2/\delta) \tag{61d}$$
$$\max\left(\lambda_0, \frac{19K^2}{n} \log(\frac{2n}{\delta})\right) \leq \lambda \leq \|\Sigma\|_{\mathcal{L}(\mathcal{F})}. \tag{61e}$$

Note that this corresponds exactly to the statement of Theorem 4.1 in the ALS setting when rescaling the constant $\delta$ by a factor 2.

**Proof of Theorem C.1:** This lemma is a straightforward adaptation of Theorem 4.1, using Lemma C.2 (and the corresponding hypotheses) instead of Lemma C.1 in order to control the term $\|P_m^\perp(\Sigma + \lambda I)^{1/2}\|_{\mathcal{L}(\mathcal{F})}$ in (56). The only difference (beyond the sampling scheme) is that the bound holds only with probability $1 - 2\delta$.

## D  PROOF OF Proposition 4.1

We prove here Proposition 4.1 which provides a sufficient condition for Assumption 2 to hold, and we state below a direct corollary of Theorem 4.1 under this condition.

**Proof of Proposition 4.1:** Take $\mu \in \mathcal{S}^\kappa$ and set $f = \mathcal{A}(\mu)$. Using the decomposition of the covariance operator introduced at the end of Section 4.2 we have

$$\begin{aligned}
\mathcal{N}_f(\lambda) &= \langle f, (\Sigma + \lambda I)^{-1} f \rangle_\mathcal{F} \\
&= \sum_{\ell \in \mathbb{N}} \frac{\langle f, e_\ell \rangle_\mathcal{F}^2}{\sigma_\ell + \lambda} \stackrel{(i)}{=} \sum_{\ell \in \mathcal{J}} \frac{\langle f, e_\ell \rangle_\mathcal{F}^2}{\sigma_\ell + \lambda} \\
&= \sum_{\ell \in \mathcal{J}} \frac{\langle f, e_\ell \rangle_\mathcal{F}^2}{\sigma_\ell^{2s}} \frac{\sigma_\ell^{2s}}{\sigma_\ell + \lambda} = \sum_{\ell \in \mathcal{J}} \frac{\langle f, e_\ell \rangle_\mathcal{F}^2}{\sigma_\ell^{2s}} \frac{\lambda^{2s}(\sigma_\ell/\lambda)^{2s}}{\lambda(\sigma_\ell/\lambda + 1)} \\
&\stackrel{(ii)}{\leq} \left( \sum_{\ell \in \mathcal{J}} \frac{\langle f, e_\ell \rangle_\mathcal{F}^2}{\sigma_\ell^{2s}} \right) \lambda^{2s-1} \\
&\leq \frac{\gamma_s}{\lambda^{1-2s}}
\end{aligned} \qquad (62)$$

where $(i)$ follows from (22a), $(ii)$ follows from the fact that $x^{2s}/(x+1) \leq 1$ for every $x > 0$ given that $2s \leq 1$, and the last inequality follows from (22b). For any $t > (3^{1-2s}\gamma_s)^{1/(2s)}$, combining (62) with the definition of $\lambda$ given in the proposition we get

$$\mathcal{N}_f(\lambda) \leq \frac{\gamma_s}{\lambda^{1-2s}} \leq t.$$

As this holds for any $\mu \in \mathcal{S}^\kappa$, using the definition (13) of $\lambda_t$ we get that $\lambda_t \leq \lambda$ i.e. (19a) is satisfied. Finally, by definition of $t$ we have $t^{2s} > 3^{1-2s}\gamma_s$ and thus

$$3\lambda t = 3\gamma_s^{\frac{1}{1-2s}} t^{-\frac{1}{1-2s}} t = \left( \frac{3^{1-2s}\gamma_s}{t^{2s}} \right)^{\frac{1}{1-2s}} < 1$$

which gives (19b). The claim about the LRIP constant is clear.

**Corollary D.1:** Using the notations and under the hypotheses of both Theorem 4.1 and proposition 4.1, for any $\delta > 0$ and $t > (3^{1-2s}\gamma_s)^{1/(2s)}$ we get with probability at least $1 - \delta$

$$\mathrm{ER}(\pi, \hat{h}) \leq \inf_{\pi_\mathfrak{S} \in \mathfrak{S}} d_{C'}(\pi_\mathfrak{S}, \pi) + 4C' \|\mathcal{A}(\pi - \pi_n)\|_\mathcal{F} \qquad (63)$$

$$\text{where} \quad C' = \frac{C_\mathcal{F}}{\sqrt{1 - 3\gamma_s^{\frac{1}{1-2s}} t^{\frac{2s}{1-2s}}}}$$

and provided that

$$m \geq \max\left( 67, 5K^2 \left(\frac{\gamma_s}{t}\right)^{-\frac{1}{1-2s}} \right) \log \frac{4K^2}{\left(\frac{\gamma_s}{t}\right)^{\frac{1}{1-2s}} \delta}. \qquad (64)$$

**Proof of Corollary D.1:** This is a direct consequence of Theorem 4.1, choosing $t, \lambda$ as given in Proposition 4.1. The bound on $m$ leverages the fact that $\mathcal{N}_\infty(\lambda) \leq K^2/\lambda$ for any $\lambda > 0$.

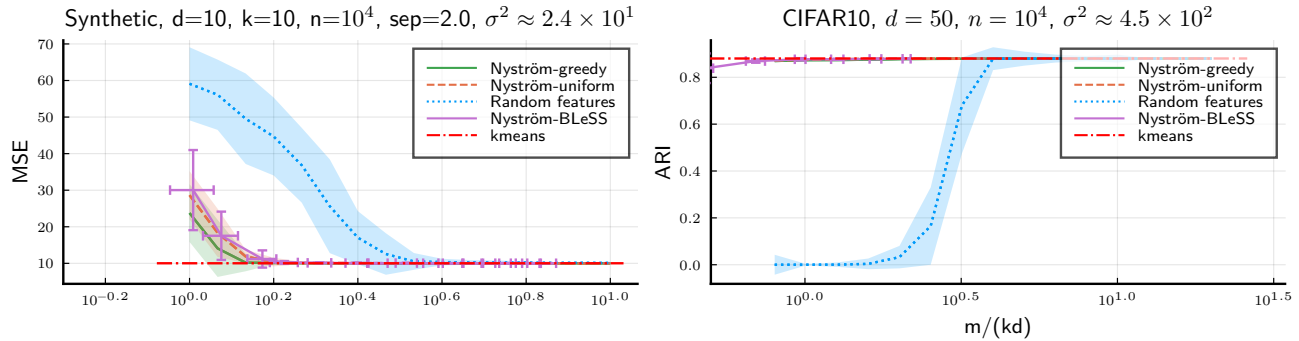

Figure 3: Median and standard deviation of the risk for a synthetic dataset and the adjusted Rand index for CIFAR10.

## E  EXPERIMENTAL RESULTS

### E.1  Datasets description

Synthetic data is drawn according to a Gaussian mixture with probability density function $\pi(x) = \frac{1}{k}\sum_{i=1}^{k} \mathcal{N}(x; \boldsymbol{\mu}_i, \mathbf{I})$ with $\boldsymbol{\mu}_i \sim \mathcal{N}(0, \sigma_{\text{inter}}^2 \mathbf{I})$ and $\sigma_{\text{inter}} = s k^{1/d}$, where $s$ is a parameter controling the separation between clusters and fixed to $s = 2.0$ unless otherwise specified. The number of samples is fixed to $n = 10^4$ or $n = 10^5$ and specified directly in the figures. The real datasets consists in vectorial features extracted from the FMA (Defferrard et al. 2016), MNIST (LeCun et al. 1998) and CIFAR10[5] datasets. FMA consists of audio features. We used the raw dataset but kept only the dimensions corresponding to the MFCC features, yielding $n = 106574$ samples in dimension $d = 20$. The MNIST data consists of $n = 70000$ handwritten digits features with $k = 10$ classes. Distorted variants are generated, and dense SIFT descriptors are extracted and used to form a $k$-nearest neighbors matrix. Spectral features are then computed by taking the $k = 10$ eigenvectors associated to the smallest positive eigenvalues of the corresponding Laplacian matrix. For CIFAR10, we use the test set to produce convolutional features before the last average pooling layer of a trained ResNet18 (He et al. 2016a,b). The network is trained on the training set of CIFAR10 for 200 epochs with SGD with momentum 0.9, learning rate 0.1 decreased by a factor 10 at epoch 100 and 150, batch-size 128, weight-decay $10^{-4}$. The extracted features are then reduced to dimension 50 with linear PCA. For each experiment we report median and standard deviation over 50 trials unless stated otherwise.

### E.2  Setup for Figure 1

The choice of the kernel variance is known to have a strong influence on the results, especially for Gaussian modeling. In order to avoid confusion, we thus manually choose a good variance for each setting rather than learning an optimal parameter automatically.

The kernel variance was fixed for clustering experiments to $\sigma^2 = 81$ for the synthetic dataset, $\sigma^2 = 5000$ for FMA, $\sigma^2 = 0.3$ for MNIST, $\sigma^2 = 450$ for CIFAR10, and for Gaussian modeling to $\sigma^2 = 24$ for the synthetic dataset, $\sigma^2 = 5000$ for FMA, $\sigma^2 = 0.095$ for Nyström and $\sigma^2 = 0.3$ for RF for MNIST, $\sigma^2 = 300$ for Nyström and $\sigma^2 = 10^4$ for RF for CIFAR10.

### E.3  Additional experiments

We provide in Figure 3 two additional plots for clustering. One is a synthetic dataset with different parameters than in the paper, and the second one corresponds to the same experiments as the one depicted in Figure 1 but shows the adjusted Rand index of the recovered clustering using the ground truth classes (rather than the MSE).

In Figure 4, we present the results for Gaussian modeling (same results than in the main paper + 1 extra synthetic setting) but also plot the variation of the risk as a function of the kernel variance. We also represent

---

[5] https://www.cs.toronto.edu/~kriz/cifar.html

**Mean Nyström Embeddings for Adaptive Compressive Learning**with a vertical line the manually chosen kernel variances, which for some of the datasets depends on the chosen approximation method. We observe than for all datasets except MNIST, Nyström approximation is more stable with respect to the choice of $\sigma^2$.

Antoine Chatalic, Luigi Carratino, Ernesto De Vito, Lorenzo Rosasco

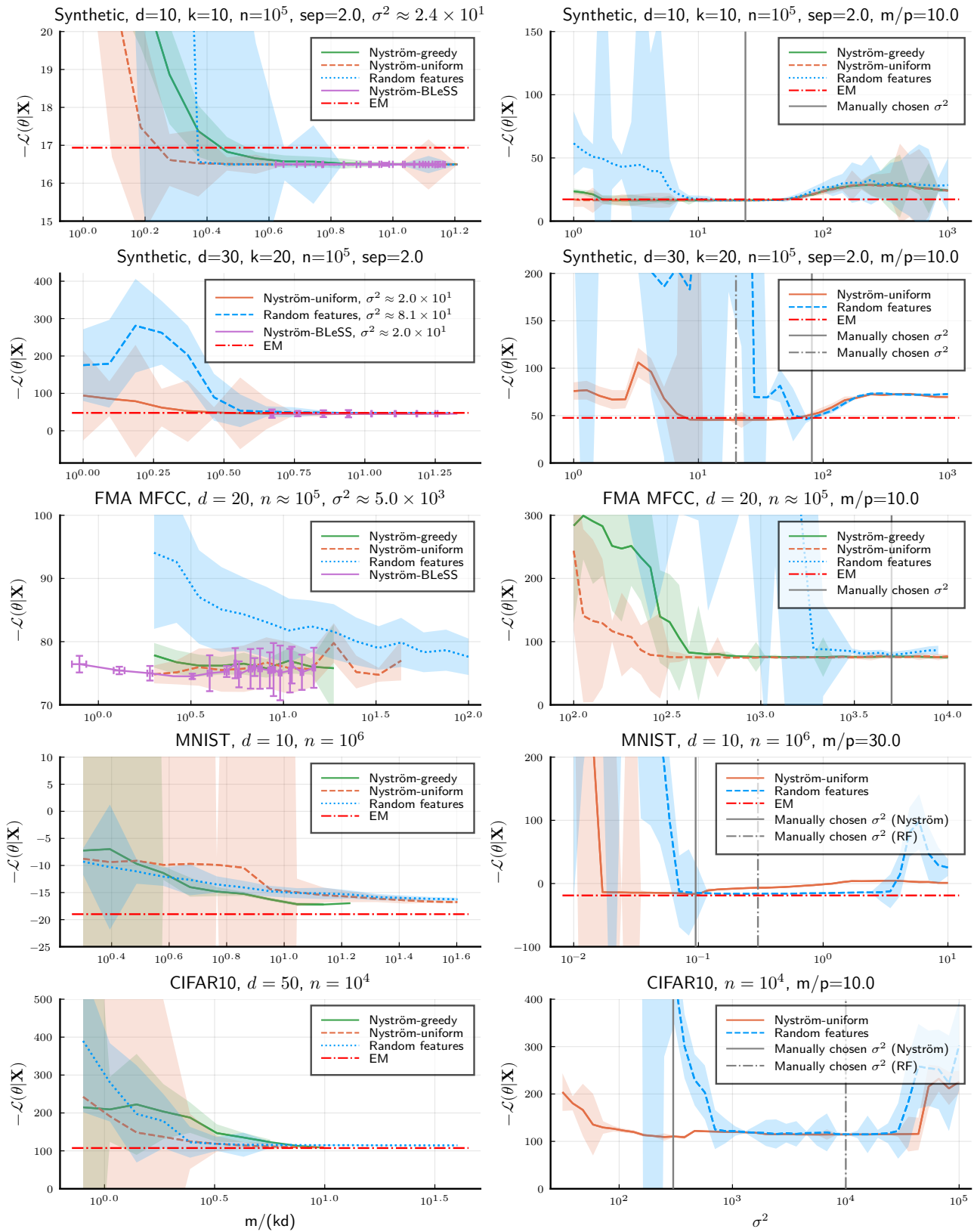

Figure 4: Median and std of the risk vs sketch size (left) and kernel variance (right) for Gaussian modeling.